\documentclass[a4paper]{article}

\usepackage{INTERSPEECH_v2}
\usepackage{amssymb,amsmath,epsfig,bm,multirow}

\title{Embedding-Based Speaker Adaptive Training of Deep Neural Networks}
\name{Xiaodong Cui, Vaibhava Goel, George Saon}
\address{IBM T. J. Watson Research Center, Yorktown Heights, NY 10598, USA}
\email{\{cuix,vgoel,gsaon\}@us.ibm.com}

\begin{document}

\maketitle
\begin{abstract}
An embedding-based speaker adaptive training (SAT) approach is proposed and investigated in this paper for deep neural network acoustic modeling. In this approach, speaker embedding vectors, which are a constant given a particular speaker, are mapped through a control network to layer-dependent element-wise affine transformations to canonicalize the internal feature representations at the output of hidden layers of a main network. The control network for generating the speaker-dependent mappings is jointly estimated with the main network for the overall speaker adaptive acoustic modeling. Experiments on large vocabulary continuous speech recognition (LVCSR) tasks show that the proposed SAT scheme can yield superior performance over the widely-used speaker-aware training using i-vectors with speaker-adapted input features.
\end{abstract}


\section{Introduction}
\label{sec:intro}

Pattern variation caused by speaker variability is one of the fundamental issues in acoustic modeling for automatic speech recognition (ASR). Such variation can give rise to covariate shift \cite{Shimodaira_Covshift} that degrades the recognition performance. Speaker adaptation \cite{Woodland_MLLR}\cite{Lee_MAP} and speaker adaptive training (SAT) \cite{Anastasakos_SAT}\cite{Gales_SAT} have been a common practice in ASR to reduce the covariate shift incurred by speaker variability.

In deep neural network (DNN) acoustic modeling, given the large number of parameters in the network and its lack of a generative structure for parameter tying, speaker adaption is typically carried out on a selected subset of parameters under the constraint of data sparsity. For instance, a common way of adaptation is to retrain or fine-tune the input or output layer of the network for a new speaker at the test time. Or, alternatively, the input or output layer is adapted by an affine transformation estimated using the adaptation data \cite{Yao_SADNN}\cite{Liao_SADNN}\cite{Sim_SADNN}. To reduce the number of parameters to be adapted given the sparse adaptation data, the output layer is further factorized in \cite{Xue_SVDSADNN}. In another effort under the name of learning hidden unit contributions (LHUC), only the amplitudes of the outputs of hidden units are re-scaled with the adaptation data \cite{Swietojanski_LHUC}.

Speaker-aware training is also widely used for adaptation of DNN acoustic models, notably with i-vectors. In \cite{Saon_IVECT}, an i-vector is estimated for each speaker and appended to all the input features from the same speaker, which is performed for both training and test speakers. DNNs learned from the features with such speaker indicators are supposed to be more speaker invariant.

SAT as another treatment of speaker variability, which is distinct from the above-mentioned adaptation techniques, aims at creating speaker invariance by explicitly introducing a speaker-dependent transformation in the feature space for canonicalization. SAT features have been a popular choice for both GMM-HMM and DNN-HMM acoustic modeling in some of the state-of-the-art LVCSR systems \cite{Saon_2016IBMSWB}. In \cite{Swietojanski_SATLHUC}, a SAT-LHUC setup is introduced where the speaker-dependent amplitude functions are directly integrated into the objective function and jointly trained with speaker independent representations.

Given the increasing complexity of DNNs, which imposes a severe problem of data sparsity, and also intrinsically lack of a generative structure to explore, canonicalizing the feature space to remove the speaker variability during training appears to be appealing. In this paper we focus on SAT of DNNs. Particularly, we propose an embedding-based SAT scheme for speaker invariance in DNN acoustic modeling. Speaker embedding maps a speaker with his/her speech data to a low-dimensional continuous vector with a fixed length. In that sense, i-vector \cite{Dehak_ivec} is one way of embedding speakers. It renders a compact representation of the speaker's acoustic characteristics. We map i-vectors to element-wise affine transformations through a control network. Since each speaker is embedded into a constant i-vector, the resulting affine transformations are speaker dependent. The affine transformations are applied to the outputs of the hidden units of selected layers of a main network for speaker normalization. The control and main networks are then jointly optimized to create canonical internal feature spaces. Experiments conducted on Babel and Switchboard LVCSR tasks show that this approach can improve over i-vector-based speaker-aware training with speaker-adapted features.

The remainder of the paper is organized as follows. Section \ref{sec:form} gives the formulation of the proposed SAT approach. Section \ref{sec:exp} provides the experimental results on Babel and Switchboard data sets followed by a discussion in Section \ref{sec:dis}. Section \ref{sec:sum} concludes the paper with a summary and speculation on the future work.

\section{Embedding-based SAT}
\label{sec:form}

SAT has been a very effective technique in ASR, especially in GMM-HMM acoustic modeling \cite{Anastasakos_SAT}\cite{Gales_SAT}. It is also commonly used in DNN-HMM acoustic modeling for speaker-adapted features \cite{Saon_2016IBMSWB} via FMLLR\cite{Gales_CMLLR}. It creates a canonical feature space in terms of speaker variability. Since neural network acoustic models with a deep structure render a hierarchical feature representation, it is intriguing to see if such canonical internal feature spaces between layers are also helpful.

Suppose there are $n_{l}$ hidden units in layer $l$ whose outputs form an $n_{l}$-dimensional internal feature space. Suppose features $\bm{x}^{(s)}$ are features from speaker $s$ in this space, $\bm{x}^{(s)} \in \mathbb{R}^{n_{l}}$. We define a speaker-dependent affine transformation on the features from speaker $s$:
\begin{align}
  \hat{\bm{x}}^{(s)} = \mathbf{A}^{(s)}_{l}\bm{x}^{(s)}+\mathbf{b}^{(s)}_{l}  \label{eqn:sat_matrix}
\end{align}
where $\mathbf{A}^{(s)}_{l}$ is a diagonal matrix
\begin{align}
    \mathbf{A}^{(s)}_{l} = \textbf{diag}\big\{a^{(s)}_{l,1},\cdots,a^{(s)}_{l,n_{l}}\big\}
\end{align}
In other words, the affine transformation is element-wise.

Define the following two vectors
\begin{align}
   & \mathbf{a}^{(s)}_{l} = [a^{(s)}_{l,1},\cdots,a^{(s)}_{l,n_{l}}]  \\
   & \mathbf{b}^{(s)}_{l} = [b^{(s)}_{l,1},\cdots,b^{(s)}_{l,n_{l}}]
\end{align}
The affine transformation can be written as
\begin{align}
     \hat{\bm{x}}^{(s)} = \mathbf{a}^{(s)}_{l} \odot \bm{x}^{(s)} \oplus \mathbf{b}^{(s)}_{l}  \label{eqn:sat_scalar}
\end{align}
where $\odot$ stands for element-wise multiplication and $\oplus$ for element-wise addition.

Furthermore, we define a mapping between the embedding vector of speaker $s$, $\bm{e}^{(s)}$, and the corresponding affine transformation
\begin{align}
   & \mathbf{a}^{(s)}_{l} = f_{la}(\bm{e}^{(s)})  \\
   & \mathbf{b}^{(s)}_{l} = f_{lb}(\bm{e}^{(s)})
\end{align}
where the mappings $f_{la}$ and $f_{lb}$ are realized by neural networks with $\bm{e}^{(s)}$ as input.

\begin{figure}[htb]
   \centering
   \centerline{\epsfig{figure=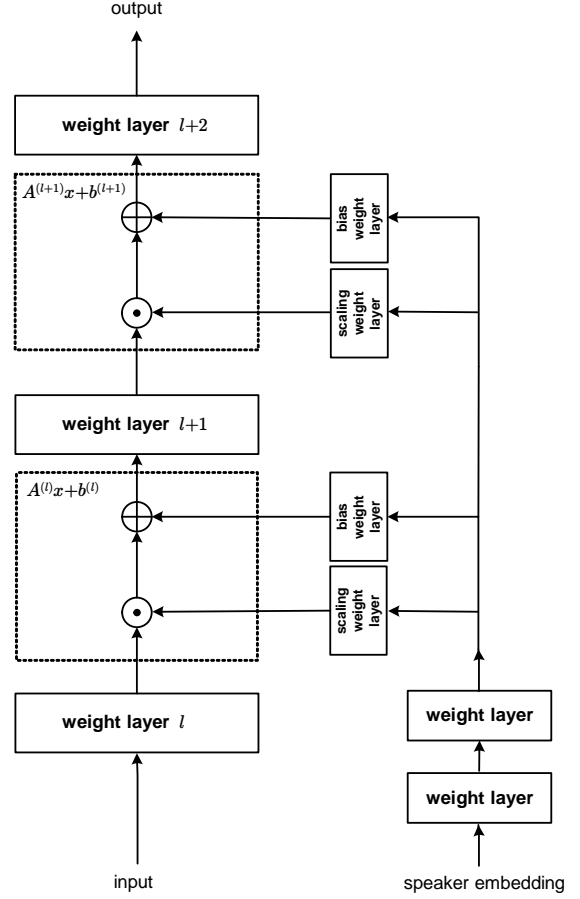, width=7.3cm, height=12cm}}
   \caption{\label{fig:satdnn}An illustration of the realization of the embedding-based speaker adaptive training with a control network (right) and a main network (left). The control network maps a speaker embedding vector to layer-dependent element-wise affine transformations. The main and control networks are jointly trained to simultaneously estimate the speaker-specific transformations and acoustic models with canonical internal feature spaces. In the figure, $\odot$ stands for element-wise multiplication while $\oplus$ for element-wise addition.}
\end{figure}

Fig.\ref{fig:satdnn} illustrates a realization of the proposed SAT scheme. The speaker embedding vectors are used as input to the network on the right which is referred to as the control network. The control network has multiple outputs which are the scaling vector $\mathbf{a}_{l}$ and bias vector $\mathbf{b}_{l}$ of the element-wise affine transformation applied to a selected number of layers $l$ of a generic network on the left referred to as the main network. The resulting multiple transformations or the scaling and bias vectors of the same transformation share the bottom layers of the control network and only split up at the last layer, denoted scaling and bias weight layers in the figure. The scaling and bias weight layers have the same number of hidden units as that of the layer to be normalized in the main network. Let $\tilde{\bm{e}}$ denote the output of the topmost layer of the shared bottom sub-network before splitting, the outputs of the scaling and bias layers for layer $l$  are given by
\begin{align}
    & \mathbf{a}_{l} = \sigma (\mathbf{W}_{la}\tilde{\bm{e}}+\mathbf{b}_{la})   \\
    & \mathbf{b}_{l} = \text{tanh} (\mathbf{W}_{lb}\tilde{\bm{e}}+\mathbf{b}_{lb})
\end{align}
where $\{\mathbf{W}_{la}, \mathbf{b}_{la}\}$ and $\{\mathbf{W}_{lb}, \mathbf{b}_{lb}\}$ are the weights and biases in the scaling and bias layers respectively and the nonlinearity for the scaling layer is sigmoid while hyperbolic tangent for the bias layer. The sigmoid nonlinearity enforces a positive scaling while the hyperbolic tangent nonlinearity allows the bias to be both positive and negative.  The configuration of the control network has been empirically evaluated and determined.

The main network represents a feedforward neural network (FNN), convolutional neural network (CNN) or recurrent neural network (RNN) used for acoustic modeling in a general sense. A number of its layers are selected to get normalized by speaker-dependent transformations for canonical internal feature spaces, which is equivalent to inserting a SAT layer right after one selected layer. In training the control and main networks are treated as one single network. The SAT transformations and the acoustic model with canonical internal feature spaces are therefore jointly optimized.

\section{Experimental results}
\label{sec:exp}

Experiments are carried out on Babel and Switchboard data sets to evaluate the proposed SAT scheme.

\subsection{Babel}
The conversational speech data in Swahili and Georgian full language pack (FLP) under the IARPA Babel program \cite{Babel} are used for training. They are the surprise languages in the Babel option period 2 and 3 evaluations respectively. The training set of each of the two languages consists of 40 hours of speech data. There are 525 training speakers in Swahili and 518 in Georgian. Word error rates (WERs) are measured on the development sets of the two languages. There are 20 hours of audio data from 142 speakers in Swahili and 20 hours from 124 speaker in Georgian.

Table \ref{tab:babel} shows a variety of DNN models and their performance on the two languages.

The first row presents a DNN model whose input is composed of 9 frames of 40-dimensional speaker-adapted FMLLR features, denoted ``fmllr'' in the table. The dimensionality of input is 360. The DNN consists of 5 hidden layers. Each hidden layer has 1,024 hidden units. The bottom three hidden layers use ReLU activation functions while the top two hidden layers use sigmoid activation functions. The softmax output layer has 3,000 units corresponding to context-dependent (CD) HMM states.

The DNN in the second row has the same network configuration except that a 100-dimensional i-vector is appended to the previous 360-dimensional input features, denoted ``fmllr+ivec'' in the table. So its input dimensionality of 460.

The models in the last two rows are the implementations of proposed SAT scheme with the i-vector-augmented FMLLR input features. In their configurations, as the control network introduces extra parameters, the topmost sigmoid hidden layer is removed. Therefore, the main network consists of 4 hidden layers -- three ReLU layers at the bottom plus one sigmoid layer on top. A SAT layer generated by the control network is inserted right after each hidden layer, which gives rise to 4 SAT layers in total. The input to the control network is the 100-dimensional i-vector, the same i-vector appended to the FMLLR features as input to the main network. The scaling and bias weight layers at the end of the control network (Fig.\ref{fig:satdnn}) have the same number of hidden units as the corresponding layer to be normalized in the main network, which in this case is 1,024. As mentioned in Section \ref{sec:form}, the control network has shared layers at the bottom before branching out for each SAT transformation. In the experiments, the numbers of hidden units in the shared layers are configured to be [100, 256, 512, 1024] with 100 being the input i-vector dimensionality. Observations are made that such a configuration that gradually maps a low-dimensional space to a high-dimensional space gives good performance. The difference between the two models in the last two rows is that the bias compensation in the SAT transformation is removed ($\mathbf{b}_{l}$ is nulled Eq.\ref{eqn:sat_matrix} or Eq.\ref{eqn:sat_scalar}) in the models presented in the 3rd row. Under this condition, it amounts to a gating mechanism with positive scaling.

All models in the table are trained under the cross-entropy (CE) criterion with randomly initialized weights. In SAT training, the control and main networks are jointly optimized. The CE training uses a mini-batch based stochastic gradient descent (SGD) algorithm with frame randomization for 20 epoches. The batch size is 256. Bigram language models (LMs) are used in decoding. The vocabulary size of Swahili is 24K and Georgian 34K.

As can be observed from the table, the speaker-aware training by appending i-vectors to FMLLR features (fmllr+ivec) gives 0.3\% absolute improvement on Swahili but does not help in Georgian. The embedding-based SAT training improves WERs on top of speaker-aware training with FMLLR features on both languages (50.4\% $\rightarrow$ 49.9\% in Swahili and 53.8\% $\rightarrow$ 52.7\% in Georgian).

In terms of SAT training, gating with only scaling the hidden unit outputs in the main network gives 0.2\% absolute improvement on Swahili and 0.7\% on Georgian. Furthermore, a full affine transformation with both scaling and bias compensation appears to be more helpful in this case. Specifically, a full SAT transformation obtains another 0.3\% absolute improvement over gating on Swahili and 0.4\% on Georgian.

\begin{table}[htb]
\caption{WERs(\%) on Swahili and Georgian with and without embedding-based SAT under the CE criterion. The model in the first row is a baseline DNN with speaker-adapted FMLLR features (denoted fmllr). The model in the second row is speaker-aware training by appending i-vectors to the fmllr features (denoted fmllr+ivec). The models in the last two rows are SAT models using fmllr+ivec features.}
\label{tab:babel}
\centering
\begin{tabular}{l|c|c} \hline
                          &    \multicolumn{2}{c}{WER}       \\ \cline{2-3}
                          &     Swahili    &   Georgian      \\ \hline\hline
  fmllr                   &     50.7       &    53.7         \\ \hline
  fmllr+ivec              &     50.4       &    53.8         \\ \hline
  fmllr+ivec+gating       &     50.2       &    53.1         \\ \hline
  fmllr+ivec+sat          &  \textbf{49.9} & \textbf{52.7}   \\ \hline
\end{tabular}
\end{table}

\subsection{Switchboard 300}

We also evaluated the embedding-based SAT on the 300-hour switchboard data set. The training set consists of 262 hours of Switchboard 1 audio with transcripts provided by the Mississippi State University. The test set is the Hub5 2000 evaluation set composed of two parts: One is the 2.1-hour switchboard (SWB) data from 40 speakers and the other is the 1.6-hour callhome (CH) data from 40 speakers.

Acoustic models are long short-term memory (LSTM) networks. We evaluated LSTMs up to 5 bi-directional layers in the main network. Each layer contains 1,024 cells with 512 on each direction. On top of the LSTM layers, there is a linear bottleneck layer with 256 hidden units followed by a softmax output layer with 32K units corresponding to CD HMM states. The LSTMs are unrolled 21 frames and trained with non-overlapping feature subsequences of that length. In SAT LSTM, the scaling and bias weight layers at the end of the control network have 1,024 hidden units which are equal in number to that of the internal state of the bi-directional layer. The numbers of hidden units in the shared layers are configured to be [100, 512, 1024] with 100 being the input i-vector dimensionality. The FMLLR features have a dimensionality of 40.

Tables \ref{tab:hub5CE} shows LSTMs trained under the CE criterion with randomly initialized weights for 25 epochs. Again the control and main networks are jointly optimized in the SAT training. Mini-batch based SGD is used for optimizing with a batch size of 128. Each batch contains subsequences of 21 frames from different utterances. The language model is a legacy model that had been used for previous publications \cite{Saon_2016IBMSWB}.It consists of a 4M 4-gram language model with a vocabulary of 31K words.

Analogous to the Babel experiments, we compare the embedding-based SAT to conventional LSTMs using FMLLR features and also speaker-aware training by appending i-vectors to FMLLR features. The SAT affine transformations are applied to the outputs of bottom one or two LSTM layers depending on the total numbers of LSTM layers. Specifically, two SAT layers are applied for the 5-layer LSTMs. Otherwise, only one SAT layer is applied to the bottom LSTM layer.

Table \ref{tab:hub5CE} shows the performance of LSTMs with different settings on SWB and CH respectively. As can be observed from the table, increasing the number of LSTM layers results in better performance until plateaued after 4 layers for LSTMs without SAT. Speaker-aware training by appending i-vectors to FMLLR features can significantly improve the performance. In a 4-layer LSTM, it reduces the WER by 0.7\% (11.4\% $\rightarrow$ 10.7\%) absolute for SWB and 1.0\% for CH (19.7\% $\rightarrow$ 18.7\%). The embedding-based SAT training can further improve the performance on top of the i-vector-based speaker-aware training. It can also improve performance for LSTMs with 5 layers. In this case, inserting two SAT layers in a 5-layer LSTM obtained 10.3\% WER in SWB and 18.4\% WER in CH compared to the speaker-aware training's 10.7\% and 18.8\% in SWB and CH, respectively.

\begin{table}[htb]
\caption{WERs(\%) on Hub5 2000 Switchboard (SWB) and Callhome (CH) of LSTMs with and without embedding-based SAT under the CE criterion using 300 hours training data. The models in the first row uses speaker-adapted FMLLRs feature (denoted fmllr). The models in the second row are speaker-aware training by appending i-vectors to the FMLLR features (denoted fmllr+ivec). The models in the third row are SAT models using fmllr+ivec features. The SAT normalization is applied after each of the bottom two LSTM layers for the 5-layer LSTM case, otherwise after the bottom LSTM layer only.}
\label{tab:hub5CE}
\centering
\begin{tabular}{l||c|c|c|c|c} \hline
                          &    \multicolumn{5}{c}{WER (SWB)}            \\ \cline{2-6}
                          &   1L   &   2L   &   3L   &   4L    &   5L   \\ \hline\hline
  fmllr                   &  13.3  &  11.9  &  11.5  &  11.4   &   --     \\ \hline
  fmllr+ivec              &  12.1  &  11.2  &  10.8  &  10.7   &  10.7     \\ \hline
  fmllr+ivec+sat          & \textbf{12.0}  & \textbf{10.9}  & \textbf{10.5}  & \textbf{10.5} &   \textbf{10.3}  \\ \hline
\end{tabular}

\bigskip

\begin{tabular}{l||c|c|c|c|c} \hline
                          &    \multicolumn{5}{c}{WER (CH)}       \\ \cline{2-6}
                          &  1L    &   2L   &   3L   &  4L    &   5L  \\ \hline\hline
  fmllr                   &  22.6  &  20.7  &  19.9  &  19.7  &   --   \\ \hline
  fmllr+ivec              &  21.3  &  19.6  &  18.9  &  18.7  &  18.8  \\ \hline
  fmllr+ivec+sat          & \textbf{21.0}  &  \textbf{19.3}  &  \textbf{18.3}  & \textbf{18.4} &  \textbf{18.4}   \\ \hline
\end{tabular}
\end{table}

Table \ref{tab:hub5ST} shows the performance of 5-layer LSTM models under the state-level Minimum Bayes Risk (sMBR) sequence training criterion. As a reference, in the first row, the CE WERs are carried over from Table \ref{tab:hub5CE}, which correspond to the last column of the table. Compared to the CE training, the sMBR-based sequence training yields 0.4\% absolute improvement on SWB and 0.3\% on CH in the baseline LSTM setup without SAT. The embedding-based SAT gives rise to additional 0.1\% absolute improvement on SWB and 0.5\% on CH after sequence training.

\begin{table}[htb]
\caption{WERs(\%) on Hub5 2000 Switchboard (SWB) and Callhome (CH) of LSTMs with and without embedding-based SAT under the sMBR sequence training criterion using 300 hours training data. The LSTM models have 5 layers with fmllr+ivec feature input.  The SAT normalization is applied after each of the bottom two LSTM layers.}
\label{tab:hub5ST}
\centering
\begin{tabular}{l|c|c} \hline
                             &    \multicolumn{2}{c}{WER}         \\ \cline{2-3}
                             &     SWB         &     CH           \\ \hline\hline
  fmllr+ivec / CE            &     10.7        &    18.8          \\ \hline
  fmllr+ivec / sMBR ST       &     10.3        &    18.5          \\ \hline
  fmllr+ivec+sat / sMBR ST   &  \textbf{10.2}  &  \textbf{18.0}   \\ \hline
\end{tabular}
\end{table}

\subsection{Switchboard 2000}

We further evaluated the embedding-based SAT on the 2000-hour switchboard data set. The training set consists of 1975 hours of segmented audio. Specifically, it contains 262 hours from Switchboard 1 data collection, 1698 hours from the Fisher data collection and 15 hours of CallHome audio.

The LSTM acoustic models are configured the same way as those used in the Switchboard 300 hours case except that the number of bi-directional LSTM layers is 6.  The feature input to the LSTM acoustic models is a fusion of FMLLR, i-Vector and logMel with its delta and double delta. Embedding-based SAT is applied to the bottom 3 LSTM layers.

The language model is also rebuilt using publicly available (e.g. LDC) training data, including Switchboard, Fisher, Gigaword, and Broadcast News and Conversations. The most relevant data is the transcripts of the 1975 hour audio data used for training the acoustic model, consisting of about 24M words. The vocabulary size is also increased from 31K to 85K words. The final LM used for decoding has 36M 4-grams.

Table \ref{tab:hub5_swb2000} presents the WERs of LSTM acoustic models with and without embedding-based SAT trained under CE and sMBR sequence training criteria, respectively. Particularly, the 7.2\% WER on SWB and 12.7\% WER on CH given by the baseline LSTM after the sequence training so far is our best single system performance without rescoring using more advanced LMs (e.g. Model M and LSTM LMs \cite{Saon_2017IBMSWB}). From the table, the embedding-based SAT yields 0.6\% absolute improvement on SWB and 0.5\% on CH after CE training. After the sequence training, it gives 0.1\% and 0.2\% absolute improvement on SWB and CH, respectively, over our best single system.

\begin{table}[htb]
\caption{WERs(\%) on Hub5 2000 Switchboard (SWB) and Callhome (CH) of LSTMs with and without embedding-based SAT under the cross-entropy (CE) and state-level Minimum Bayes Risk (sMBR) criteria using 2,000 hours training data. The LSTM models have 6 layers with fmllr+ivec+logmel+$\Delta$+$\Delta^{2}$ feature input.  The SAT normalization is applied after each of the bottom three LSTM layers.}
\label{tab:hub5_swb2000}
\centering
\begin{tabular}{l|c|c} \hline
                                                            &    \multicolumn{2}{c}{WER}         \\ \cline{2-3}
                                                            &     SWB         &     CH           \\ \hline\hline
  fmllr+ivec+(logmel+$\Delta$+$\Delta^{2}$) / CE            &     8.1         &    13.5          \\ \hline
  fmllr+ivec+(logmel+$\Delta$+$\Delta^{2}$) / sMBR ST       &     7.2         &    12.7          \\ \hline\hline
    fmllr+ivec+(logmel+$\Delta$+$\Delta^{2}$)+sat / CE        &   7.5         &    13.0          \\ \hline
  fmllr+ivec+(logmel+$\Delta$+$\Delta^{2}$)+sat / sMBR ST   &  \textbf{7.1}   &  \textbf{12.5}   \\ \hline
\end{tabular}
\end{table}

\section{Related Work and Discussion}
\label{sec:dis}

The embedding-based SAT training maps speaker embedding vectors (i-vectors) to speaker-dependent affine transformations to normalize internal feature spaces of a DNN acoustic model. In \cite{Miao_SATDNN}, i-vectors are mapped to speaker-dependent biases which are added to input features to compensate the speaker variability. The speaker normalization is only conducted in the input feature space. In that sense, it still belongs to the family of feature space normalization at input of DNNs. In \cite{Samarakoon_FHL}, factorized hidden layers (FHL) are constructed as a linear combination of speaker dependent bases for neural network adaptation. The bases are initialized with i-vectors and optimized during training. The FHL strategy is conceptually different from the embedding-based SAT investigated in this work.

The embedding-based SAT introduces full affine transformations with scaling and bias. If the bias is nulled, it amounts to a gating mechanism that performs element-wise positive scaling of the outputs of the hidden units, which is equivalent to a SAT-LHUC form with the scaling vectors constructed from speaker embedding vectors.

The embedding vectors provide a flexible way of generating transformations with various granularity. For instance, the embedding can be performed on utterances, conversations or speakers and the resulting transformations will have the corresponding granularity.

Speech signals are known to be affected by a variety of variabilities, most notably speaker, environment and channel. Creating canonical representation spaces by removing these speech variabilities has always been pursued in acoustic modeling. In the embedding-based adaptive training framework, if one can estimate the relevant embedding of a known variability, the generated transformations will canonicalize the internal feature spaces of a DNN acoustic model accordingly. It is fairly straightforward to extend this framework for noise or channel normalization if the necessary  embedding can be accomplished.

\section{Summary and future work}
\label{sec:sum}

In this paper we proposed an embedding-based speaker adaptive training approach for acoustic modeling using deep neural networks. In this approach, i-vectors are mapped to layer-dependent element-wise affine transformations for internal feature space canonicalization. The mapping is realized through a control network and it is jointly trained with a main network the outputs of whose selected hidden layers are normalized. The approach is evaluated on Babel and Switchboard data sets. It shows that this approach can improve performance over DNNs with speaker-adapted input features and also speaker-aware training using i-vectors.

Looking forward, we will investigate the optimization of the SAT models under the sequence training criterion. Besides, alternative ways of speaker embedding other than i-vectors will be examined.

\bibliographystyle{IEEEtran}

\bibliography{embedding_satdnn}

\begin{thebibliography}{10}
\providecommand{\url}[1]{#1}
\csname url@samestyle\endcsname
\providecommand{\newblock}{\relax}
\providecommand{\bibinfo}[2]{#2}
\providecommand{\BIBentrySTDinterwordspacing}{\spaceskip=0pt\relax}
\providecommand{\BIBentryALTinterwordstretchfactor}{4}
\providecommand{\BIBentryALTinterwordspacing}{\spaceskip=\fontdimen2\font plus
\BIBentryALTinterwordstretchfactor\fontdimen3\font minus
  \fontdimen4\font\relax}
\providecommand{\BIBforeignlanguage}[2]{{%
\expandafter\ifx\csname l@#1\endcsname\relax
\typeout{** WARNING: IEEEtran.bst: No hyphenation pattern has been}%
\typeout{** loaded for the language `#1'. Using the pattern for}%
\typeout{** the default language instead.}%
\else
\language=\csname l@#1\endcsname
\fi
#2}}
\providecommand{\BIBdecl}{\relax}
\BIBdecl

\bibitem{Shimodaira_Covshift}
H.~Shimodaira, ``Improving predictive inference under covariate shift by
  weighting the log-likelihood function,'' \emph{Journal of Statistical
  Planning and Inference}, vol.~90, no.~2, pp. 227--244, 2000.

\bibitem{Woodland_MLLR}
C.~J. Leggetter and P.~C. Woodland, ``Maximum likelihood linear regression for
  speaker adaptation of continuous density hidden {M}arkov models,''
  \emph{Computer Speech and Language}, vol.~9, pp. 171--185, 1995.

\bibitem{Lee_MAP}
J.~L. Gauvain and C.~H. Lee, ``Maximum a posteriori estimation for multivariate
  gaussian mixture observations of markov chains,'' \emph{IEEE Transactions on
  Speech and Audio Processing,}, vol.~2, no.~2, pp. 291--298, 1994.

\bibitem{Anastasakos_SAT}
T.~Anastasakos, J.~McDonough, R.~Schwartz, and J.~Makhoul, ``A compact model
  for speaker-adaptive training,'' in \emph{International Conference on Spoken
  Language Processing (ICSLP)}, 1996, pp. 1137--1140.

\bibitem{Gales_SAT}
M.~J.~F. Gales, ``Cluster adaptive training of hidden {M}arkov models,''
  \emph{IEEE Transactions on Speech and Audio Processing}, vol.~8, no.~4, pp.
  417--428, 2000.

\bibitem{Yao_SADNN}
K.~Yao, D.~Yu, F.~Seide, H.~Su, L.~Deng, and Y.~Gong, ``Adaptation of
  context-dependent deep neural networks for automatic speech recognition,'' in
  \emph{Spoken Language Technology Workshop (SLT)}, 2012, pp. 366--369.

\bibitem{Liao_SADNN}
H.~Liao, ``Speaker adaptation of context dependent deep neural networks,'' in
  \emph{International Conference on Acoustics, Speech and Signal Processing
  (ICASSP)}, 2013, pp. 7947--7951.

\bibitem{Sim_SADNN}
B.~Li and K.~C. Sim, ``Comparison of discriminative input and output
  transformations for speaker adaptation in the hybrid {NN/HMM} systems,'' in
  \emph{Interspeech}, 2010, pp. 526--529.

\bibitem{Xue_SVDSADNN}
J.~Xue, J.~Li, D.~Yu, M.~Seltzer, and Y.~Gong, ``Singular value decomposition
  based low-footprint speaker adaptation and personalization for deep neural
  network,'' in \emph{International Conference on Acoustics, Speech and Signal
  Processing (ICASSP)}, 2014, pp. 6359--6363.

\bibitem{Swietojanski_LHUC}
P.~Swietojanski and S.~Renals, ``Learning hidden unit contributions for
  unsupervised speaker adaptation of neural network acoustic models,'' in
  \emph{Spoken Language Technology Workshop (SLT)}, 2014, pp. 171--176.

\bibitem{Saon_IVECT}
G.~Saon, H.~Soltau, D.~Nahamoo, and M.~Picheny, ``Speaker adaptation of neural
  network acoustic models using {I}-vectors,'' in \emph{Automatic Speech
  Recognition and Understanding Workshop (ASRU)}, 2013, pp. 55--59.

\bibitem{Saon_2016IBMSWB}
G.~Saon, T.~Sercu, S.~Rennie, and H.-K. Kuo, ``The {IBM} 2016 {E}nglish
  conversational telephone speech recognition system,'' in \emph{Interspeech},
  2016, pp. 7--11.

\bibitem{Swietojanski_SATLHUC}
P.~Swietojanski and S.~Renals, ``{SAT-LHUC}: speaker adaptive training for
  learning hidden unit contributions,'' in \emph{International Conference on
  Acoustics, Speech and Signal Processing (ICASSP)}, 2016, pp. 5010--5014.

\bibitem{Dehak_ivec}
N.~Dehak, P.~Kenny, R.~Dehak, P.~Dumouchel, and P.~Ouellet, ``Front-end factor
  analysis for speaker verification,'' \emph{IEEE Transactions on Audio,
  Speech, and Language Processing,}, vol.~19, no.~4, pp. 788--798, 2011.

\bibitem{Gales_CMLLR}
M.~J.~F. Gales, ``Maximum likelihood linear transformations for {HMM}-based
  speech recognition,'' \emph{Computer Speech and Language}, vol.~12, pp.
  75--98, 1998.

\bibitem{Babel}
Https://www.iarpa.gov/index.php/research-programs/babel.

\bibitem{Saon_2017IBMSWB}
G.~Saon, G.~Kurata, T.~Sercu, K.~Audhkhasi, S.~Thomas, D.~Dimitriadis, X.~Cui,
  B.~Ramabhadran, M.~Picheny, L.-L. Lim, B.~Roomi, and P.~Hall, ``English
  conversational telephone speech recognition by humans and machines,'' in
  \emph{Interspeech}, 2017, pp. 132--136.

\bibitem{Miao_SATDNN}
Y.~Miao, H.~Zhang, and F.~Metze, ``Speaker adaptive training of deep neural
  network acoustic models using {I}-vectors,'' \emph{IEEE/ACM Transactions on
  Audio, Speech, and Language Processing}, vol.~23, no.~11, pp. 1938--1949,
  2015.

\bibitem{Samarakoon_FHL}
L.~Samarakoon and K.~C. Sim, ``Factorized hidden layer adaptation for deep
  neural network based acoustic modeling,'' \emph{IEEE/ACM Transactions on
  Audio, Speech, and Language Processing}, vol.~24, no.~12, pp. 2241--2250,
  2016.

\end{thebibliography}

\end{document}